\documentclass{article}
\usepackage{arxiv}
\usepackage[utf8]{inputenc} 
\usepackage[T1]{fontenc}    
\usepackage{hyperref}       
\usepackage{url}            
\usepackage{booktabs}       
\usepackage{amsfonts}       
\usepackage{nicefrac}       
\usepackage{microtype}      
\usepackage{xcolor}         
\usepackage{multirow} 
\usepackage{graphicx}
\usepackage{amsmath}
\usepackage{bbm}

\title{medDreamer: Model-Based Reinforcement Learning with Latent Imagination on Complex EHRs for Clinical Decision Support}

\author{%
  Qianyi Xu
    \\
  National University of Singapore\\
  \texttt{xuqianyi@u.nus.edu} \\
  \And
  Gousia Habib \\
  National University of Singapore\\
  \texttt{gousiya1@nus.edu.sg} \\
  \And
  Feng Wu
    \\
  National University of Singapore\\
  \texttt{wufeng@u.nus.edu} \\
  \And
  Dilruk Perera
  \thanks{Correspondence to dilruk@nus.edu.sg} \\
  National University of Singapore\\
  \texttt{dilruk@nus.edu.sg} \\
  \And
  Mengling Feng \\
  National University of Singapore\\
  \texttt{ephfm@nus.edu.sg} \\
}

\begin{document}

\maketitle

\begin{abstract}
 Timely and personalized treatment decisions are essential across a wide range of healthcare settings where patient responses can vary significantly and evolve over time. Clinical data used to support these treatment decisions are often irregularly sampled, where missing data frequencies may implicitly convey information about the patient's condition. Existing Reinforcement Learning (RL) based clinical decision support systems often ignore the missing patterns and distort them with coarse discretization and simple imputation. They are also predominantly model-free and largely depend on retrospective data, which could lead to insufficient exploration and bias by historical behaviors. To address these limitations, we propose medDreamer, a novel model-based reinforcement learning framework for personalized treatment recommendation. medDreamer contains a world model with an Adaptive Feature Integration module that simulates latent patient states from irregular data and a two-phase policy trained on a hybrid of real and imagined trajectories. This enables learning optimal policies that go beyond the sub-optimality of historical clinical decisions, while remaining close to real clinical data. We evaluate medDreamer on both sepsis and mechanical ventilation treatment tasks using two large-scale Electronic Health Records (EHRs) datasets. Comprehensive evaluations show that medDreamer significantly outperforms model-free and model-based baselines in both clinical outcomes and off-policy metrics. 
\end{abstract}

\maketitle

\section{Introduction}
Critical decision making in healthcare often involves managing highly dynamic and uncertain patient trajectories. Conditions such as sepsis or acute respiratory failure require timely and personalized interventions. However, the high complexity and heterogeneity of physiological responses challenge standardized care protocols. RL offers a principled framework for learning adaptive treatment strategies from complex data as it mirrors the sequential clinical decision-making process and directly optimizes long-term patient outcomes. Unlike standard supervised learning algorithms that aim to mimic historical clinician actions, RL can, in theory, surpass suboptimal clinical practices by learning decision treatment policies that are better aligned with long-term outcome-driven objectives\cite{banumathi2025reinforcement}.

However, despite RL's promise and increased interest, deploying RL solutions in the real-world clinical setting remains limited. The unique characteristics of EHR data, such as irregular sampling and highly sparse observations, pose significant challenges for standard RL methods. Clinical events occur sporadically based on the patient state and clinician decisions, which leads to the fact EHR data are highly irregular and often showcase \textit{informative missingness}\cite{rubin1976inference}, where the absence of measurements conveys key clinical information about patient conditions and dynamics. Existing solutions commonly rely on coarse discretization of patient trajectories into regular time windows (e.g., 4 hourly), and erroneous imputations using heuristics such as mean or median \cite{komorowski2018artificial}. This could obscure important underlying patient dynamics, introduce information loss and modeling bias.

Additionally, unlike traditional RL methods applied in gaming or robotics \cite{goldwaser2020deep,yang2020overview}, real-world exploration during policy training is ethically infeasible, as experimentation on real patients is not an option. Therefore, offline RL is utilized in most solutions where we completely learn from retrospective data.
Model-free offline RL methods have become the de facto in clinical decision-making tasks due to their algorithmic simplicity and minimal assumptions about environment dynamics. However, these solutions are tightly coupled to the historical actions without the ability to simulate or explore alternative decision paths that can help generalize beyond the potentially suboptimal and biased past clinical actions \cite{nambiar2023deep,luo2024reinforcement, amirahmadi2023deep}. This severely hinders the widely used model-free offline RL solutions from capturing the complexity of clinical trajectories and recommending robust, outcome-improving treatments in critical healthcare environments. To overcome these challenges, a fundamentally different approach is required that can model patient dynamics more faithfully, allowing counterfactual reasoning and supporting policy learning beyond what clinicians have historically practiced, where applicable.


Accordingly, we propose medDreamer, a novel model-based RL framework designed to address the dual challenges of irregular clinical data and the constraints of model-free offline policy learning. medDreamer consists of a learned patient world model and a two-phase hybrid policy training pipeline. The world model captures patient dynamics from irregular observations and time-elapsed intervals using a novel Adaptive Feature Integration (AFI) Module, and generates simulated patient trajectories via latent imagination\cite{hafner2024masteringdiversedomainsworld, burchi2025learning, samsami2024mastering}. By modeling trajectories directly in the latent space, medDreamer avoids the inaccuracies introduced by conventional time discretization and error-prone imputations, allowing more accurate next state prediction in the sparse observational data. Importantly, medDreamer supports rich policy exploration entirely offline. By sampling trajectories from the patient world model, we enable rich policy exploration while respecting the ethical boundaries. To ensure policy safety and alignment, we introduce a two-phase training strategy. Unlike the common practice of model-based RL approaches that purely train on imagined data \cite{hafner2024masteringdiversedomainsworld}, policies are initially trained on hybrid trajectories that contain both real and imagined steps, and then fine-tuned on fully simulated trajectories. We evaluate our solution on two clinical decision-making tasks: sepsis treatment recommendation and mechanical ventilation (MV) management, both leading causes of ICU mortality where optimal treatment policies remain an open question\cite{nauka2025challenges, goligher2016clinical}. These challenges call for an effective data-driven solution capable of generalizing beyond limited retrospective clinical data to learn better treatment policies. Our main contributions are as follows: 
\begin{itemize}
    \item \textbf{A World model with AFI for irregular data handling.} We introduce a patient dynamics model with an AFI module that effectively captures complex temporal patterns in sparse and irregularly sampled EHR data and support effective latent imagination. 
    \item \textbf{A hybrid two-phase policy training framework.} We propose a two-phase policy training framework that blends real and imagined trajectories, enabling safe exploration that maintains alignment with observed clinical behavior.
    \item \textbf{Extensive evaluation across multiple clinical tasks.} We comprehensively evaluate our framework on two large real-world datasets for both sepsis treatment and mechanical ventilation management tasks, demonstrating better treatment efficacy and survival rates, with a significant mortality rate decrease of 11.19\% and 5.83\% compared to clinicians.
\end{itemize}

\section{Related Work}
Inspired by the success in domains such as robotics \cite{tang2025deep}, game playing \cite{mnih2013atari, Szita2012} and autonomous driving \cite{peng2024improving}, RL has been widely adopted for various healthcare treatment recommendation tasks \cite{ali2022reinforcement} for both chronic \cite{si_25,yu2021reinforcement} and acute \cite{jia2020safe,raghu2017deep} treatment recommendation tasks.
In this section, we review the literature from two perspectives: representation learning for patient state modeling and the evolution of model-based RL for treatment recommendation. 

\textbf{Representation learning for RL in treatment recommendation:}  Accurate patient state representation is crucial for applying RL in clinical decision support. Several works have explored differential equation-based approaches to effectively represent irregular patient dynamics. GRU-ODE-Bayes\cite{de2019gru} models sporadic EHR data using ordinary differential equations (ODEs), and subsequent work extends this for modeling treatment effects over time\cite{de2022predicting}. Others utilize controlled differential equations (CDEs)\cite{killian2020empirical, kidger2020neural, morrill2021neural, seedat2022continuous} to track how patient states evolve. However, these methods build a continuous path over irregular states and are not suitable for creating discrete time latent trajectories. These methods also focus on irregularities for a single feature across time but do not account for the correlation among different features. In contrast, our approach is able to analyze both inter-feature and intra-feature dynamics to construct robust patient representations from sparse EHR data for downstream latent dynamics modeling and policy optimization.\\
\textbf{Model-based RL for treatment recommendation.}
The majority of past solutions utilize model-free RL, where treatment policies are learned directly from retrospective EHRs without an explicit model of patient dynamics \cite{liu2024reinforcement, kondrup2023towards, estiri2024model, tu2025offline}. While efficient given enough data, model-free RL often suffers from sample inefficiency and limited capacity to reason about counterfactual outcomes since interactions with the real environment are not possible under clinical settings. 
In contrast, model-based reinforcement learning (MBRL) offers a principled framework for optimizing treatment strategies by simulating patient dynamics. Many previous works in this area build transition models based on Bayesian Neural Networks (BNNs). Killian et al.\cite{killian2017robust} and Yao et al.\cite{yao2018direct} propose HiP-MDPs to adapt transition dynamics based on latent patient embeddings for personalized decision-making. Raghu et al.\cite{raghu2018model} utilize a BNN to model transition dynamics for sepsis treatment. Parbhoo et al.\cite{parbhoo2017combining} apply Gaussian process kernels for HIV treatment optimization, and Peng et al.\cite{peng2018improving} extend similar methods to sepsis management. However, these methods operate directly in the real observation space, making it difficult to learn under noisy and irregular EHR data.
Instead, our work infers the next state in the latent space by incorporating latent imagination\cite{hafner2019dream,hafner2020mastering,hafner2024masteringdiversedomainsworld}. Further distinguishing our approach, unlike traditional model-based approaches that rely solely on world model-generated trajectories for policy learning\cite{schrittwieser2020mastering,levine2016end}, we employ a two-phased approach to first ground the policy in real trajectories before introducing simulated rollouts. Previous offline MBRL methods, such as COMBO\cite{yu2021combo} and MOPO\cite{yu2020mopo}, also use both real and simulated data uniformly during training. However, these approaches lack explicit grounding and exploration phases, and operate directly in raw observation spaces. 
Our proposed approach, medDreamer, addresses these limitations by introducing a structured, two-phased policy learning approach. We rely more on historical data during the first phase and shift to full exploration during the second phase, ensuring clinical reliability for healthcare applications.


\section{Methodology}

\subsection{Problem Formulation}
We formulate the treatment recommendation problem as a finite-horizon Markov Decision Process (MDP), defined by the tuple $(\mathcal{O}, \mathcal{S}, \mathcal{A}, \mathcal{P}, r, \gamma)$, where $\mathcal{O}$ denotes the observation space, consisting of partially observed clinical measurements $o_t \in \mathbb{R}^d$, with associated time gaps $\delta_t$ and masks $m_t\in\{0,1\}^d$, $\mathcal{S}$ is the latent patient state space inferred from observations, $\mathcal{A}$ is the action space representing treatment decisions, $\mathcal{P}$ denotes the unknown transition dynamics over latent states, $r$ is the reward function capturing clinical improvement or deterioration, and $\gamma$ the discount factor. At each timestep $t$, a patient’s state is partially observed through a tuple $(o_t, \Delta_t)$ where $o_t \in \mathbb{R}^d$ denotes observed physiological features (vitals, labs, demographics), and $\Delta_t$ is the time elapsed since the last available observation. The observation is mapped to a latent state $s_t \in \mathcal{S}$ through an encoder. The agent selects an action $a_t \in \mathcal{A}$ and transitions to the next state $s_{t+1} \sim P(s_{t+1}|s_t,a_t)$, and receives a scalar reward $r_t$. The objective is to learn a policy $\pi_\theta(a_t|s_t)$ that maximizes the expected return over horizon $T$, $\mathbb{E}_{\pi_\theta}[\Sigma_{t=1}^T\gamma^t r(s_t,a_t)]$.

\subsection{Solution Overview}
Clinical data from EHRs exhibit a challenging structure of double sparsity: (1) temporal sparsity, where observations occur at irregular time intervals, and (2) feature sparsity, where only a subset of features are recorded at each timestep. These characteristics limit the direct application of standard RL approaches and motivate our design of specialized modules for handling irregularity, missingness and complex latent dynamics. To address these challenges, we propose medDreamer, a model-based offline RL framework with two key components, (1) a \textbf{patient world model} that learns to simulate clinical trajectories through latent imagination, and (2) a \textbf{two-phase hybrid policy training schema} that enables safe counterfactual reasoning from purely offline data.
 
\begin{figure}
  \centering
  \includegraphics[width=\linewidth]{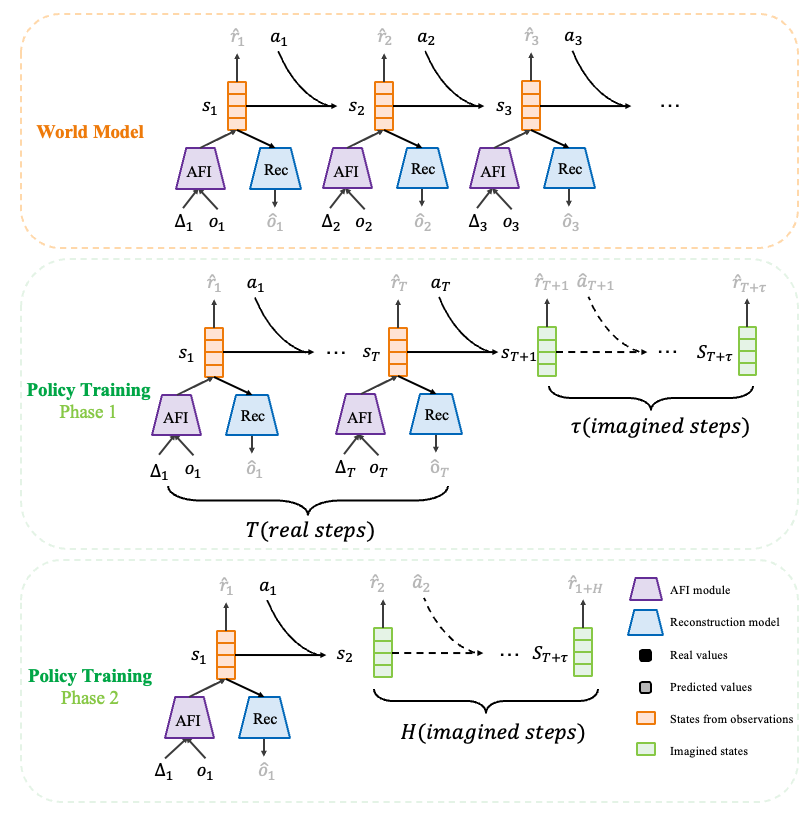} 
  \caption{Our medDreamer architecture is divided into two stages: World model training and Policy training. The world model uses time interval \(\Delta_t\), observation\(o_t\) and clinician action \(a_t\) to learn patient latent dynamics through reconstruction and predict reward \(\hat{r}_t\). Policy training consists of two phases: In Phase 1 we use \(T\) steps of states generated from real clinical decisions and imagine \(\tau\) steps into the future and train the policy using the concatenated states. In Phase 2 we use full imagined trajectories of length \(H\).}
  \label{fig:model_architecture}
\end{figure}

Our framework introduces an AFI module for processing temporally irregular, partially observed input sequences, and a specialized transition model to learn robust patient dynamics. Since jointly training the policy and model end-to-end could lead to instability in the healthcare settings, we decouple training by first training the world model, and then optimizing the policy. This allows stable policy learning and clinically meaningful generalization. The architecture is illustrated in Figure \ref{fig:model_architecture}.
Given the high variability in patient dynamics and the importance of capturing subtle physiological changes, we first train a stable world model before proceeding to policy optimization. We also design a novel two-phase training scheme for policy to enable safe, ethically compliant, and clinically grounded exploration of counterfactual scenarios. The architecture is illustrated in Figure \ref{fig:model_architecture}.

\subsection{Reward Function}
We first present the reward function used in the task. We design task-specific reward functions to reflect clinical progression and outcomes for both sepsis treatment and MV management. The total reward is composed of an \textit{intermittent reward} that captures physiological improvements or deterioration and a \textit{terminal reward} that reflects patient survival.

\paragraph{Intermittent rewards}  
For sepsis treatment, we define the intermediate reward \( R^{\text{Sep}}_{\text{im}}(s_t, s_{t+1}) \) using changes in the Sequential Organ Failure Assessment (SOFA) score and lactate levels\cite{raghu2017deep,komorowski2018artificial}: 

\begin{align}
R^{\text{Sep}}_{\text{im}}(s_t, s_{t+1}) &= 
a_{sep} \cdot \mathbbm{1}[s_{t+1}^{\text{SOFA}} = s_t^{\text{SOFA}} \land s_{t+1}^{\text{SOFA}} > 0] \notag \\
&\quad + b_{sep} \cdot (s_{t+1}^{\text{SOFA}} - s_t^{\text{SOFA}}) \notag \\
&\quad + c_{sep} \cdot \tanh(s_{t+1}^{\text{lactate}} - s_t^{\text{lactate}})
\label{eq:rim}
\end{align}
where \( s_t^{\text{SOFA}} \)is the SOFA score at time \( t \), \( s_t^{\text{lactate}} \) is the lactate level at time \( t \) and \( a_{sep}, b_{sep}, c_{sep} \) are predefined constant weights by clinicians.

\paragraph{Ventilation reward.}  
The intermediate reward encourages physiological targets to stay within clinically acceptable ranges for oxygen saturation (SpO\textsubscript{2}) and mean blood pressure (MBP). 

\begin{equation}
\begin{aligned}
R^{\text{Vent}}_{\mathrm{im}}(s_t, s_{t+1}) =\;&
a_{vent} \cdot \left( 
\mathbbm{1}_{[94 \le s_{t+1}^{\mathrm{SpO2}} \le 98]} 
- \tfrac{1}{2} \cdot 
\mathbbm{1}_{[s_{t+1}^{\mathrm{SpO2}} < 94 \,\vee\, s_{t+1}^{\mathrm{SpO2}} > 98]} 
\right) \\
+&\; b_{vent} \cdot \left(
\mathbbm{1}_{[70 \le s_{t+1}^{\mathrm{MBP}} \le 80]} 
- \tfrac{1}{2} \cdot 
\mathbbm{1}_{[s_{t+1}^{\mathrm{MBP}} < 70 \,\vee\, s_{t+1}^{\mathrm{MBP}} > 80]}
\right)
\end{aligned}
\end{equation}

where \( s_{t+1}^{\mathrm{SpO2}} \) and \( s_{t+1}^{\mathrm{MBP}} \) denote the respective SpO\textsubscript{2} and MBP at time \( t+1 \), \( a_{vent} \) and \( b_{vent} \) are constant weights. 

\paragraph{Terminal Reward.}  
To reflect patient outcomes, we apply a terminal reward \( R_{ter} \in \mathbb{R} \) for each episode end: 

\begin{equation}
R(s_t, s_{t+1}) =
\begin{cases}
+R_{ter}, & \text{if } s_{t+1} \in S_T \cap S_{\text{sur}} \\
-R_{ter}, & \text{if } s_{t+1} \in S_T \cap S_{\text{dec}} \\
R_{\text{im}}(s_t, s_{t+1}), & \text{otherwise}
\end{cases}
\label{eq:r}
\end{equation}
where $R_{ter}$ is a constant set by clinicians, \( S_T \)means set of terminal states, \( S_{\text{sur}} \) means the set of survived patients and\( S_{\text{dec}} \) for deceased patients.

\subsection{Adaptive Feature Integration (AFI)}
Clinical time series are often characterized by irregular sampling and feature sparsity. To effectively address these issues without erroneous discretization and imputations, we design an AFI module which treats missingness as a structural signal. The AFI module builds on a Factorization Machine (FM)\cite{rendle2010factorization} based architecture to explicitly model both intra-feature temporal dynamics and inter-feature interactions across variables. Specifically AFI computes higher-order interactions among these features that extract robust representations from partially observed EHR data while preserving clinically meaningful patterns that helps the downstream policy learning task.

To keep the information within missing values, we encode both the observed feature values and their temporal context into a unified input representation\cite{che2018recurrent}. Suppose that we have time series $O = (o_1,o_2,\ldots,o_{T})\in\mathbb{R}^{T\times D}$ of length $T$ and dimension $D$. At time step $t$, the mask vector is defined as,

\begin{equation}
  m_t^d = 
    \begin{cases}
      1, & \text{if } o_t^d \text{ is observed}, \\
      0, & \text{otherwise}
    \end{cases}
\end{equation}

\noindent and the time interval vector $\Delta_t^d \in\mathbb{R}$ is computed from timestamps $p_t$ using,

\begin{equation}
  \Delta^d_{t}=
  \begin{cases}
      p_t-p_{t-1}+\Delta_{t-1}^d, & t>1,m_{t-1}^d=0,\\[0.2em]
      p_t-p_{t-1},                & t>1,m_{t-1}^d=1,\\[0.2em]
      0,                          & t=1.
  \end{cases}
\end{equation}

\noindent We encode both the observed values and time intervals into vector embeddings using separate linear projections of  dimension~$k$:

\begin{equation}
  \mathbf{E}^{(o)}_t=W_oo_t\in\mathbb{R}^{D\times k},\quad
  \mathbf{E}^{(\Delta)}_t= W_\delta\Delta_t\in\mathbb{R}^{D\times k}.
\end{equation}

\noindent These are concatenated to form a joint embedding tensor \(\mathbf{E}_t = [\,\mathbf{E}^{(o)}_t \!\mid\! \mathbf{E}^{(\Delta)}_t\,] \in \mathbb{R}^{D \times 2k}\). Let \(\mathbf{e}_t^d\in\mathbb{R}^{2k}\) denote the embedding for feature $d$. We apply a FM over these embeddings to capture pairwise feature interactions as,
\begin{equation}
  \operatorname{FM}\!\bigl(\mathbf{E}_t\bigr)=\tfrac12
  \Bigl(\bigl(\textstyle\sum_{j=1}^{d}\mathbf{e}_t^d\bigr)^{\!2}-
        \textstyle\sum_{j=1}^{d}(\mathbf{e}_t^d)^{\,2}\Bigr)\in\mathbb{R}^{2k}
\end{equation}

Finally, the processed observation vector is computed as, 
\begin{equation}
  \tilde{o_t}=\sigma\!\Bigl(\operatorname{Linear}(o_t)+
                  \operatorname{FM}\!\bigl(\mathbf{E}_t\bigr)\Bigr) \in \mathbb{R}^{2k}.
\end{equation}

\subsection{Latent World Model}
To enable latent rollouts beyond observed trajectories, we learn a generative latent dynamics model that captures the evolution of patient states in the clinical environment. This model enables simulating plausible future trajectories in the latent space conditioned on treatment, without relying on future observations. 

\subsubsection{Posterior and Prior Dynamics:}
At each timestep $t$, we utilize a latent state $s_t\in \mathbb{R}^d$ that summarizes the patient history. During training, we infer this state using a posterior model as,
\begin{equation}
  s_t \sim q_\phi(s_t \mid s_{t-1}, a_{t-1}, \tilde{o_{t}}). 
  \label{eq:posterior}
\end{equation}

\noindent To generate rollouts without depending on true observations, we train a prior transition model purely based on latent states as,

\begin{equation}
  \hat{s}_t \sim p_\theta(\hat{s}_t \mid s_{t-1}, a_{t-1}).
  \label{eq:prior}
\end{equation}

\noindent Both $q_\phi$ and $p_\theta$ are parametrized as GRU network based sequential models due to its ability to capture long-term dependencies. The training objective for the transition model (prior) (see Equation \ref{eq:prior}) and representation model (posterior) (see Equation \ref{eq:posterior}) is to maximize the evidence lower bound (ELBO)\cite{jordan1999introduction}. This is achieved by minimizing the Kullback-Leibler (KL) divergence between the prior and the posterior distribution as,

\begin{equation}
\mathcal{L}_{\mathrm{D}}^{t} \; \dot{=} \;
-\text{KL} \bigl(
    p(s_t \mid s_{t-1}, a_{t-1}, \tilde{o_t})
    \;\Vert\;
    q(s_t \mid s_{t-1}, a_{t-1})
\bigr).
\end{equation}
\subsubsection{Auxiliary Prediction Heads:}
The rest of the world model consists of a reward model, continuation model, and reconstruction model. The reward model learns to predict immediate rewards from latent features, while the continuation model learns whether the trajectory ends. The reconstruction model can reconstruct observations from the latent states to make sure it contains enough information for future planning. These models are summarized as follows:
\begin{equation}
\begin{aligned}
\text{Reward model:} \quad & \hat{r}_t \sim p_\phi(\hat{r}_t \mid s_t) \\
\text{Continuation model:} \quad & \hat{c}_t \sim p_\phi(\hat{c}_t \mid s_t) \\
\text{Reconstruction model:} \quad & \hat{o}_t \sim p_\phi(\hat{o}_t \mid s_t).
\end{aligned}
\end{equation}
Loss functions for the three models are introduced as follows: $\mathcal{L}_{\text{rew}}$ uses a symlog-scaled two-hot encoding scheme, which discretizes the reward space into 255 buckets and computes a smooth approximation of log-likelihood using soft-assignment based on distance. This approach ensures stability across tasks with highly variable reward magnitudes. $\mathcal{L}_{\text{con}}$ uses logistic regression and adopts a Bernoulli log-likelihood loss modeling whether the episode continues or terminates. To enhance robustness when decoding targets with large dynamic ranges or signed magnitudes, we adopt a symlog-based squared distance loss for $\mathcal{L}_{\text{rec}}$, defined as the squared difference between the predicted value and the symlog-transformed target. We also applied masks to $\mathcal{L}_{\text{rec}}$, and their gradients of the missing values are not propagated to avoid biased updates. We jointly train the full model by minimizing the following objective:

\begin{equation}
\mathcal{L}(\phi) \doteq \mathbb{E}_{q_\phi} \Bigg[
\sum_{t=1}^{T} \Big(
\mathcal{L}_{\text{rew}}(\phi)
+ \mathcal{L}_{\text{con}}(\phi)
+ \mathcal{L}_{\text{rec}}(\phi)
+ \mathcal{L}_{\text{D}}(\phi)
\Big) \Bigg].
\end{equation}

\subsection{Policy Learning via Latent Imagination}
By leveraging the trained world model, we learn an effective treatment policy using a two-phased learning strategy. In Phase 1, we train our policy fully on real patient trajectories, while in Phase 2, the policy is refined using fully imagined future rollouts over horizons. This two-phased approach combines real-world clinical data with simulated rollouts, effectively balancing clinical grounding and exploratory learning, suitable for healthcare settings. The two phases of the learning strategy are detailed below.
The RL agent learns an initial policy from retrospective clinical data. This phase prioritizes grounding the policy in past patient histories to ensure stability and alignment with real-world clinical patterns.

\subsubsection{Phase 1, Clinically Grounded Policy Initialization via Hybrid States:}
In Phase 1, we train the actor and critic using a mixture of real patient trajectories and imagined steps. We first generate patient states of length \(T\) using real observations and historical clinical decisions, and then roll out for \(\tau\) steps into the future. Here we choose \(\tau=10\) which is much less than \(T=50\) since we want the training to prioritize grounding the policy in past patient histories to ensure stability and alignment with real-world clinical patterns. The final trajectories for policy training are the concatenation of real states and imagined states. 

Starting from the final posterior state \(\mathbf{s}_T\), we simulate future states \(\mathbf{\hat{s}}_{T+1:T+\tau}\) and actions \(\mathbf{\hat{a}}_{T+1:T+\tau}\) using the learned transition model and actor. This produces a hybrid rollout states
\begin{equation}
    S = [s_{1:T}, \hat{s}_{T+1:T+\tau}],
\end{equation}

which includes both real and imagined latent states. Rewards and continuation probabilities \((\hat{r}_t, \hat{\gamma}_t)\) are predicted by the reward and discount heads from the latent features.

We optimize the actor using the REINFORCE algorithm\cite{williams1992simple} with a learned baseline. The actor loss is defined as:
\begin{equation}
\mathcal{L}_{\text{actor}}(\theta) \doteq -\sum_{t=1}^{T+\tau} 
\operatorname{sg} \left( R_t^\lambda - V_\psi(s_t) \right) 
\log \pi_\theta(a_t \mid s_t) 
+ \eta \cdot \mathbb{H} \left[ \pi_\theta(a_t \mid s_t) \right],
\end{equation}
where \(R_t^\lambda\) is the bootstrapped \(\lambda\)-return, and \(\mathbb{H}[\cdot]\) denotes the policy entropy to encourage exploration. The stop-gradient operator \(\operatorname{sg}(\cdot)\) prevents backpropagation through the advantage.

The critic \(v_\psi\) predicts a distribution over returns. We train it using maximum likelihood on bootstrapped \(\lambda\)-returns:
\begin{equation}
\begin{aligned}
\mathcal{L}^{p1}_{\text{critic}}(\psi) \doteq\; & - \sum_{t=1}^{T+\tau} \log p_\psi\left(R_t^\lambda \mid s_t\right), \\
R_t^\lambda \doteq\; & \hat{r}_t + \gamma_t \left( (1 - \lambda) \cdot v_t + \lambda \cdot R_{t+1}^\lambda \right), \\
R_{T+\tau}^\lambda \doteq\; & v_{T+\tau},
\end{aligned}
\end{equation}
where \(v_t = \mathbb{E}[v_\psi(\cdot \mid s_t)]\) is the expected return under the critic's output distribution and \( \hat{r}_t \) denotes the imagined reward and value at step \( t \), and \( \gamma_t \) is the discount factor. The return \( R_t^\lambda \) incorporates both immediate rewards and long-term predictions, improving the robustness and foresight of the value learning process.
To stabilize value targets, we maintain a slow-moving target critic \(\bar{V}_\psi\), updated via exponential moving average with a rate of 0.02. This phase enables the agent to generalize beyond real transitions by exploring new latent paths, while still benefiting from structured grounding via the world model. Bootstrapping with predicted rewards ensures reward scale consistency between real and imagined steps.

\subsubsection{Phase 2, Strategic Policy Refinement through Long-Horizon Imagination:}
In Phase 2, the policy learned in Phase 1 is refined using fully imagined trajectories generated by the world model. This releases the full potential of exploration of the agent, enabling the policy to learn from longer planning. By simulating long-horizon dynamics, the agent can reason over extended temporal dependencies, enabling more coherent and farsighted policy improvements. Here for each state starting from \(t=0\), the model imagines \(H = 15\) future steps and is trained solely on the rollouts. 
\begin{equation}
\begin{aligned}
\mathcal{L}^{p2}_{\text{critic}}(\psi) \doteq\; & - \sum_{t=1}^{H} \log p_\psi\left(R_t^\lambda \mid s_t\right), \\
R_t^\lambda \doteq\; & \hat{r}_t + \gamma_t \left( (1 - \lambda) \cdot v_t + \lambda \cdot R_{t+1}^\lambda \right), \\
R_{H}^\lambda \doteq\; & v_{H},
\end{aligned}
\end{equation}

This stage plays a critical role in consolidating the agent’s understanding of long-term value, enabling generalization to unseen situations during deployment. Together, the two-phase training scheme enables efficient and stable policy learning. In Phase 1, the policy benefits from predominantly real states and short-horizon imagination that encourage safe and conservative learning based on accurate latent transitions. Phase 2 builds upon this foundation by leveraging long-horizon imagination to refine the policy through deep planning and foresight. This progressive strategy balances reliability and exploration, ultimately leading to a robust policy capable of handling complex, long-term decision-making tasks.

\section{Experiments}
We evaluate our method, medDreamer on two high-impact clinical decision-making tasks, namely, \textit{sepsis treatment recommendation} and \textit{mechanical ventilation management}.

\subsection{Experiment Setup}
\subsubsection{Datasets:} For sepsis treatment, we use the large-scale MIMIC-IV \cite{johnson2020mimic} database, extracting 21,233 adult septic patients. We exclude patients with missing mortality outcomes or demographics, those with treatment withdrawal or multiple ICU stays. Each trajectory starts 24 hours before diagnosis and ends at 48 hours post-diagnosis. Each observation has 40 physiological features, and the outcome is 90-day mortality. For MV management, we use the eICU Collaborative Research (eICU)\cite{pollard2018eicu} database, and extract patients aged \(\geq\)16 who received MV for at least 24 hours. We also exclude patients with missing mortality outcomes or demographics, those ventilated for \(>\)14 days, and those with missing interventions. After filtering, 21,595 patients were retained with 41 physiological features, and the outcome is hospital mortality. See Appendix section\ref{sec:feature} for full lists of features and actions.

\subsubsection{Baselines:} We evaluate medDreamer against a range of model-based and model-free RL baselines on both tasks. We take reference from current benchmarks for treatment recommendation \cite{luo2024dtr,luo2024reinforcement,hargrave2024epicare} and included naive baseline \textit{Random} for sanity check, and a supervised learning \textit{Clinician} Policy. For model-free baselines, we selected the ones that frequently appear in the benchmarks including Double Deep Q-Network (DDQN) \cite{raghu2017deep}, EZ-Vent \cite{liu2024reinforcement} based on Batch-Constrained Q-Learning (BCQ) \cite{fujimoto2019off} and DeepVent \cite{kondrup2023towards} based on Conservative Q-Learning (CQL) \cite{kumar2020conservative}. For model-based baselines, we choose a representative world model via latent imagination DreamerV3 \cite{hafner2024masteringdiversedomainsworld} which is the latest version of Dreamer framework\cite{hafner2019dream}, \textit{MBRL-BNN} \cite{raghu2018model} which uses a BNN to model patient dynamics for sepsis treatment and \textit{MBRL-GMM} \cite{li2022electronic} based Gaussian Mixture Model (GMM) and creates a surrogate world model using retrospective data for glucose control in DKA patients.

\begin{table*}[th!]
\small
\centering
\caption{OPE metrics and estimated mortality rates (with 95\% CI) for sepsis treatment and MV management on MIMIC and eICU datasets.}
\resizebox{\textwidth}{!}{
\begin{tabular}{lccccc|ccccc}
\toprule
\multirow{2}{*}{\textbf{Model}} 
& \multicolumn{5}{c|}{\textbf{Sepsis}} 
& \multicolumn{5}{c}{\textbf{MV}} \\
\cmidrule(lr){2-6} \cmidrule(lr){7-11}
& \textbf{WIS}$\uparrow$ & \textbf{WPDIS}$\uparrow$ & \textbf{CWPDIS}$\uparrow$ & \textbf{Mortality(\%)}$\downarrow$ & \textbf{95\% CI} 
& \textbf{WIS}$\uparrow$ & \textbf{WPDIS}$\uparrow$ & \textbf{CWPDIS}$\uparrow$& \textbf{Mortality(\%)}$\downarrow$ & \textbf{95\% CI} \\
\midrule
Random & 2.17 & -0.01 & -0.01 & -- & -- & 9.93 & -0.02 & 0.09 & -- & --\\
Clinician & 4.87 & 0.14 & 0.09 & 23.44 & [22.12, 24.77] & 13.31 & 0.24 & 0.18 & 18.27 & [15.48, 21.06] \\
\midrule
EZ-Vent\cite{liu2024reinforcement}(BCQ)      & 10.13 & 0.02 & 0.07 & 18.47 & [18.45, 18.49] & 7.09 & -0.10 & 0.07 &  13.78 & [13.76,  13.8]\\
DeepVent\cite{kondrup2023towards}(CQL)       & \textbf{10.42} & 0.03 & 0.07 & 19.39 & [19.38, 19.39] & 7.71 & -0.10 & 0.08 & 13.8 & [13.79  13.8]\\
DDQN\cite{raghu2017deep}                & 10.10 & 0.02 & 0.07 & 17.77 & [17.75, 17.79]  & 7.85 & -0.10 & 0.08 & 21.5 & [21.43, 21.56] \\
\midrule
MBRL-BNN\cite{raghu2018model}  & 4.90 & 0.00 & 0.01 & 16.87  & [14.5, 18.53]   & 
6.34 & 0.03 & 0.08 & 16.7 & [13.77, 19.62] \\
MBRL-GMM\cite{li2022electronic}  & -0.74 & -0.01 & -0.01 & 17.53 & [17.51, 17.55]& 
4.09 & -0.05 & 0.04 & 16.8 & [16.79, 16.81] \\
Dreamer\cite{hafner2024masteringdiversedomainsworld} & 6.85 & 0.20 & 0.13 & 15.37 & [13.13, 17.61] & 12.95 & 0.28 & 0.20 & 13.96 & [13.91, 16.57]\\

\midrule
\textbf{medDreamer}  & 9.25 & \textbf{0.27} & \textbf{0.19} & \textbf{14.25} & [14.6, 17.5] & \textbf{24.06} & \textbf{0.32} & \textbf{0.25} & \textbf{12.44} & [11.42, 13.45] \\
\bottomrule
\end{tabular}
}
\label{tab:ope_mimic_eicu}
\end{table*}

\subsubsection{Evaluation Metrics:} Since we are unable to collect feedback from real patients, we evaluate our policy using multiple off-policy evaluation (OPE) metrics based on importance sampling \cite{tokdar2010importance} that estimates expectations under trained policy using samples from the behavior policy, including WIS, WPDIS, CWPDIS and ESS \cite{thomas2015safe}. We also plot the mortality against expected return \cite{wang2018supervised} when executing clinicians' actions. This plot helps us to infer the estimated mortality by matching the average return of recommended actions. We calculate the difference between dosages recommended by RL policy and clinician policy and plot it against the mortality. A good policy should be a v-shape, indicating that the mortality is lower when clinicians agree with our RL agent and higher otherwise.

\subsubsection{Training Details and Hyperparameters:} All experiments are conducted on a single NVIDIA Tesla V100 GPU. For world model training, we use a batch size of 64, batch length of 50. We imagine 10 steps for phase 1 and 15 steps for phase 2. All models use the same decoder, reward predictor, continue predictor, and actor critic based on MLP with 2 hidden layers of size 512 with layer norm and a SiLU activation. The dynamics model uses a GRU cell with a 256-dimensional hidden state and a stochastic latent space of size 32. 
A learned initial hidden state is used to initialize each episode. The learning rate is $1 \times 10^{-4}$ for world model and phase 1 and $1 \times 10^{-5}$ for phase 2 policy training.

\subsection{Results and Discussion}

In this section, we aim to answer the following key questions.

\subsubsection{\textbf{RQ1: How well does our policy perform under offline evaluation and mortality estimation?}} 
Table\ref{tab:ope_mimic_eicu} presents the OPE metrics and estimated mortality rates for all models across the sepsis and MV tasks. Our proposed medDreamer consistently performs competitively across key metrics and achieves the lowest estimated mortality on both tasks. For the sepsis task, medDreamer obtains the highest scores in WPDIS and CWPDIS, as well as the lowest estimated mortality rate of 14.25\%, with a narrow 95\% confidence interval [14.6, 15.7], suggesting reliable estimation. While model-free baselines such as EZ-Vent and DeepVent achieve stronger WIS scores, WIS is known to be sensitive to large importance ratios and can exhibit high variance. Thus, a lower WIS score does not necessarily indicate inferior performance, especially when not considered alongside weighted or cumulatively weighted variants. In contrast, CWPDIS is typically more stable and indicative of generalization beyond the behavior policy, where medDreamer demonstrates superior values. For the MV task, medDreamer again outperforms all baselines in WIS, WPDIS, CWPDIS, and achieves the lowest estimated mortality rate of 12.44\% [11.42, 13.45], outperforming model-free methods and Dreamer. These results suggest that medDreamer's policy not only generalizes better but also recommends actions associated with improved clinical outcomes under evaluation assumptions. We include a random policy as a lower bound sanity check, representing random decisions made without any knowledge of patient states. It demonstrates that MBRL-GMM is not learning a meaningful world model by simply matching data pairs that exist in the historical dataset. We find that while model-based methods generally promote better exploration, their performance under OPE is often constrained by the accuracy of the learned dynamics model and the degree of alignment with the behavior policy. Dreamer shows improved performance over prior MBRL baselines, possibly due to its use of latent imagination rather than direct modeling in the raw observation space. This further highlights that directly modeling irregular, high-frequency data in a structured latent space can better capture patient dynamics.

\subsubsection{\textbf{RQ2: How does the learned policy align/differ from clinician decisions in treatment strategies?}}
\begin{figure}[htbp]
  \centering
  \includegraphics[width=0.85\linewidth]{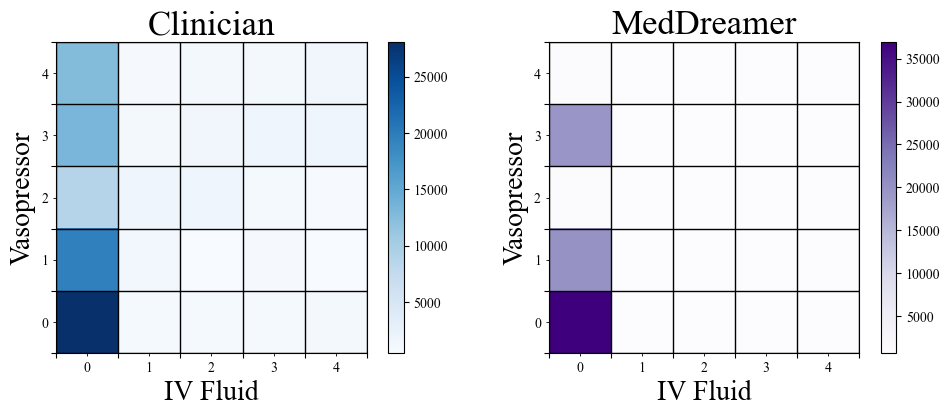} 
  \caption{Differences between clinicians' actions and policy actions for sepsis task.}
  \label{fig:action_distribution_mimic}
\end{figure}

\begin{figure}[htbp]
  \centering
  \includegraphics[width=0.95\linewidth]{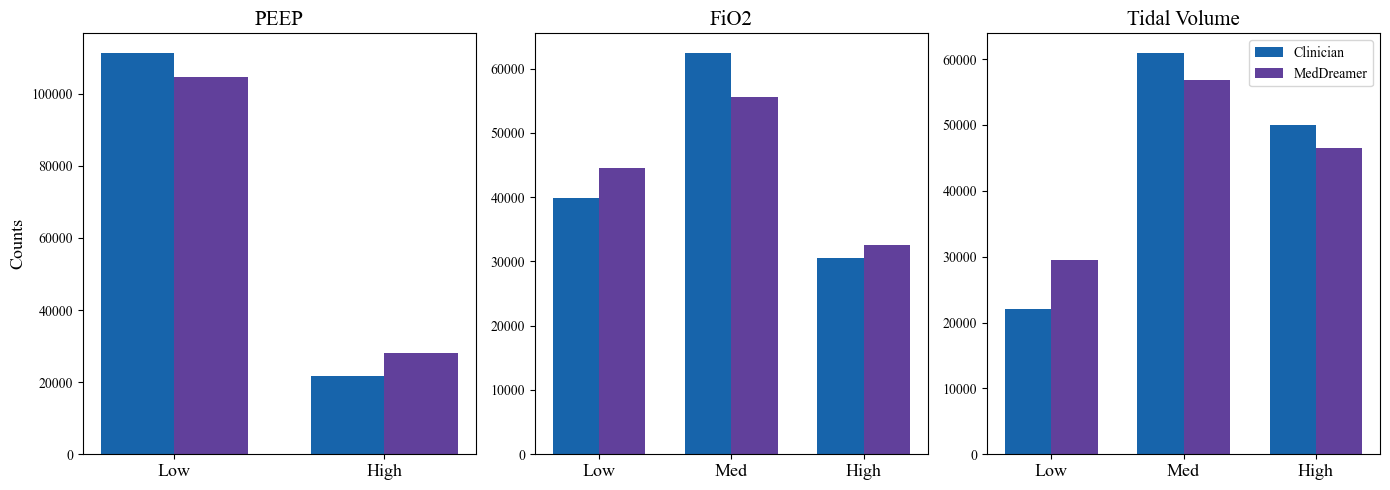} 
  \caption{Differences between clinicians' actions and policy actions for MV task.}
  \label{fig:action_distribution_eicu}
\end{figure}

We compare the action distributions of medDreamer and clinicians across both sepsis treatment (Figure\ref{fig:action_distribution_mimic}) and ventilator settings (Figure\ref{fig:action_distribution_eicu}). As shown in the joint action heatmaps, clinicians predominantly favored minimal IV fluid and vasopressor use, indicating a conservative treatment strategy. In contrast, medDreamer exhibits a broader distribution, particularly with increased use of moderate vasopressor levels, suggesting a tendency toward more proactive hemodynamic intervention. Additionally, in terms of ventilator settings, medDreamer generally adapts lower FiO\textsubscript{2} and tidal volumes compared to clinicians, aligning with lung-protective and oxygen-conservative strategies. While both approaches favor low PEEP settings, medDreamer shows a slightly higher usage of elevated PEEP. 

\begin{figure}[htbp]
  \centering
  \includegraphics[width=\linewidth]{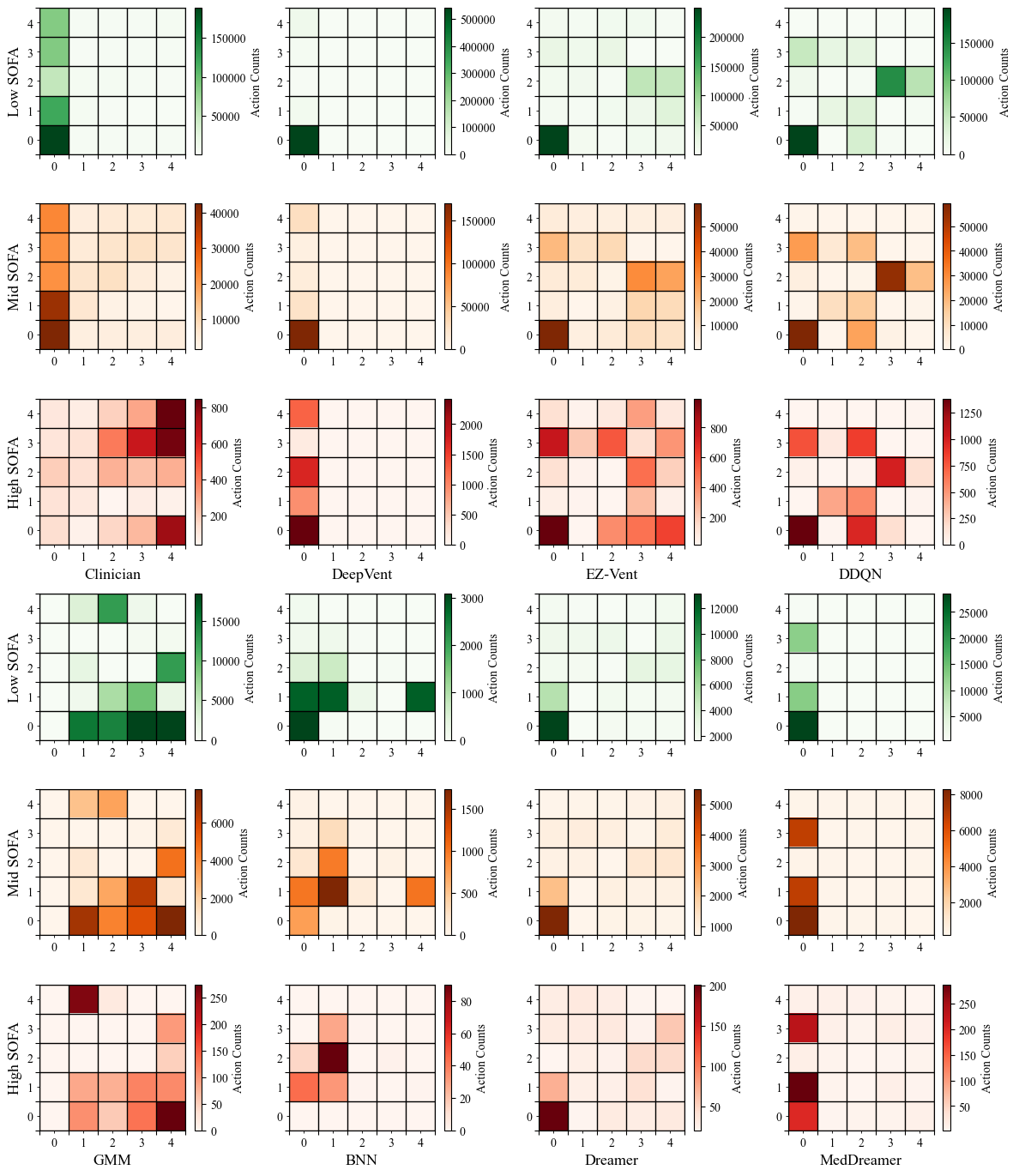} 
  \caption{Treatment action distributions for sepsis task.}
  \label{fig:action_distribution_full}
\end{figure}

The SOFA score is widely used in sepsis care to assess organ dysfunction, where higher scores indicate increased mortality risk. We observe the policy behavior across varying clinical severity by categorizing the patient into 4 SOFA ranges: Low (\(\leq 5\)), Mid (6-14), High (\(\geq 15\)), and All. We compare sepsis action distributions across SOFA severity levels to assess treatment adaptation among baseline models (see Figure \ref{fig:action_distribution_full}). Clinicians displayed clear progression from conservative to aggressive interventions with increasing severity. Our method generally follows this trend, for example, in its increased use of vasopressors under high SOFA conditions. However, medDreamer is highly conservative in the use of IV fluid, providing a new perspective on decreasing IV fluid dosage. In contrast, model-free baselines like DDQN and EZ-Vent exhibit more diffuse and less interpretable action patterns, suggesting potential instability or lack of calibration. Other model-based methods, such as MBRL-GMM and MBRL-BNN, favor consistent low-level actions but lack sufficient flexibility to adjust to patient severity. These results demonstrate medDreamer's ability to learn severity-aware treatment strategies that align more closely with clinical patterns.

\begin{figure}
  \centering
  \includegraphics[width=\linewidth]{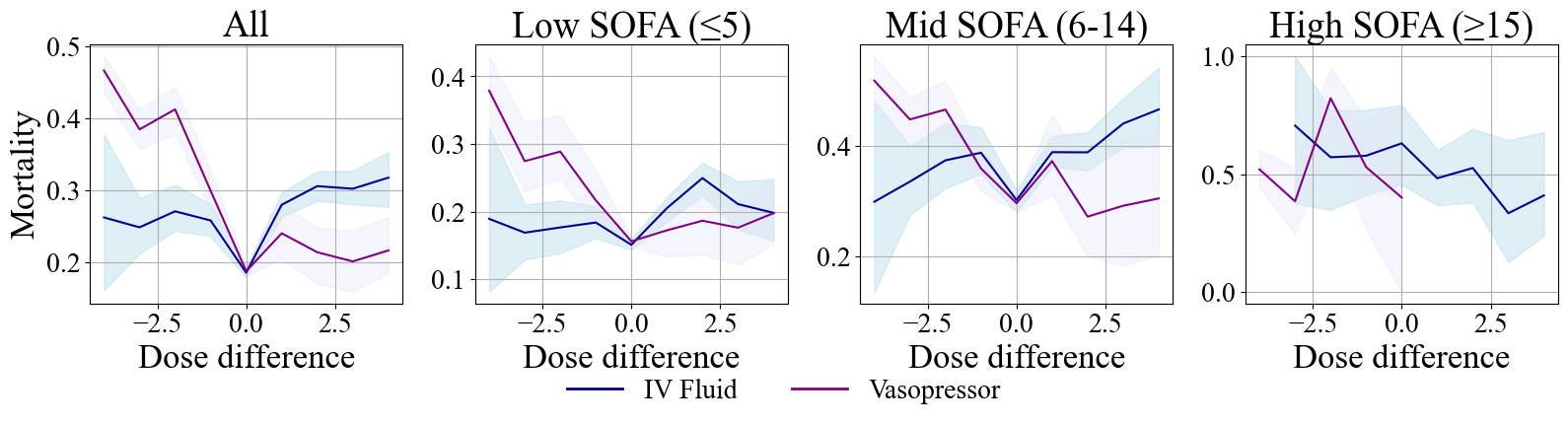} 
  \caption{Doses recommended by policy minus clinician vs mortality stratified by SOFA. For patients with high SOFA, there are no cases where policy recommends higher dose of Vasopressor than clinicians.}
  \label{fig:v_plot}
\end{figure}

Figure. \ref{fig:v_plot} shows the difference in recommended dosages between the policy and clinicians, and corresponding mortality. Except for high SOFA, all groups show a distinct V-shaped pattern, where mortality is lowest when the policy aligns with clinician decisions (i.e., x=0) and increases in divergence when in either direction. High SOFA exhibits less consistent patterns, reflecting increased complexity and variability of treating severe patients, where clinical decisions are more uncertain. This may be because current practices in high-acuity sepsis settings are known to be suboptimal.

\subsubsection{\textbf{RQ3: Is the learned value function predictive of patient survival?}}
\begin{figure}[h!]
    \centering
    \includegraphics[width=\linewidth]{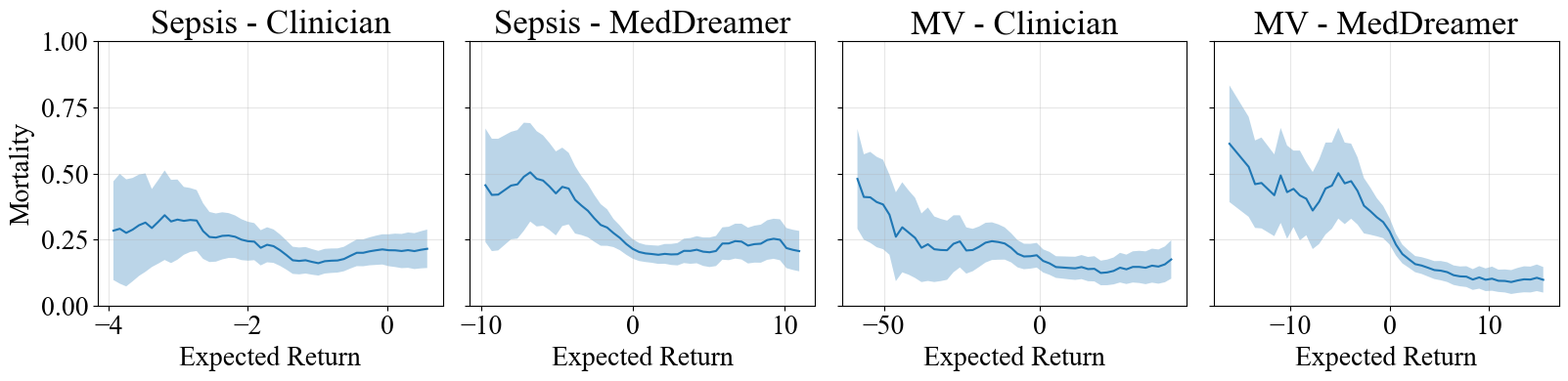} 
    \caption{Mortality vs expected return showing medDreamer's ability to assign lower returns to higher risk.} 
    \label{fig:mortality_vs_return} 
\end{figure}

To further assess the quality of learned value functions, we also visualize the relationship between expected return and observed mortality in Figure \ref{fig:mortality_vs_return}. For both sepsis and MV tasks, medDreamer demonstrates a clearer negative correlation between expected return and mortality compared to the clinician baseline, indicating better risk stratification. Please refer to Appendix \ref{sec:mort_vs_return} for the plots of other baselines.

\begin{table*}[h!]
\small
\centering
\caption{Ablations of medDreamer for sepsis treatment.}
\begin{tabular}{lccc|ccc}
\toprule
\multirow{2}{*}{\textbf{Model}} 
& \multicolumn{3}{c|}{\textbf{Sepsis}} 
& \multicolumn{3}{c}{\textbf{MV}} \\
\cmidrule(lr){2-4} \cmidrule(lr){5-7}
& \textbf{WIS}$\uparrow$ & \textbf{WPDIS}$\uparrow$ & \textbf{CWPDIS}$\uparrow$ 
& \textbf{WIS}$\uparrow$ & \textbf{WPDIS}$\uparrow$ & \textbf{CWPDIS}$\uparrow$ \\
\midrule
No AFI & 9.10 & 0.20 & 0.14 & 15.77 & 0.29 & 0.18 \\
MDP1 Only & 7.78 & 0.19 & 0.14 & 17.11 & 0.29 & 0.22\\
MDP2 Only & 6.81 & 0.21 & 0.13 & 14.27 & 0.31 & 0.22\\
E2E & 10.11 & 0.21 & 0.13 & 8.22 & 0.26 & 0.16 \\
\midrule
ActorCritic & 7.65 & 0.23 &	0.17 & 19.81 & 0.31 & 0.23\\
Curriculum & 7.65 & 0.23 & 0.17 & 15.93 & 0.29 & 0.21\\
\midrule
medDreamer($\tau=5$) & 8.34 & 0.26 & 0.19 & 21.99 & 0.29 & 0.17\\
\textbf{MedDreamer($\tau=10$)} & \textbf{9.24} & \textbf{0.27} & \textbf{0.19} & \textbf{24.06} & \textbf{0.32} & \textbf{0.25} \\
medDreamer($\tau=30$) & 8.81 & 0.21 & 0.15 & 8.21 & 0.29 & 0.18\\
medDreamer($\tau=50$) & 8.42 & 0.24 & 0.17 & 18.21 & 0.28 & 0.17\\
\midrule
medDreamer($H=5$) &  8.66 & 0.24 & 0.17 & 16.91 & 0.29 & 0.2\\
\textbf{MedDreamer($H=15$)} & \textbf{9.24} & \textbf{0.27} & \textbf{0.19} & \textbf{24.06} & \textbf{0.32} & \textbf{0.25} \\
medDreamer($H=25$) & 9.08 & 0.26 & 0.19 & 16.92 & 0.28 & 0.21\\
medDreamer($H=35$) & 9.01 & 0.24 & 0.17 & 16.15 & 0.28 & 0.2\\
\midrule
LSTM & 8.89 & 0.22 & 0.16 & 16.58 & 0.29 & 0.19\\
Transformer & 9.11 & 0.27 & 0.19 & 16.90 & 0.29 & 0.20 \\
Random mask & 9.17 & 0.25 & 0.19 & 18.16 & 0.26 & 0.16\\ 
\bottomrule
\end{tabular}
\label{tab:ablation}
\end{table*}
\subsubsection{\textbf{RQ4: How do architectural and imagination design choices affect policy performance?}}

For comprehensive validation, we conducted three ablation studies (see Table \ref{tab:ablation}) as follows.

(1) \textit{Module and algorithm design validity}: We examine different components of our algorithm design including AFI module and two-phase training. To assess how efficient our model is in handling missing and irregular EHR data, we compare medDreamer to a normal encoder instead of using AFI, where missing values are imputed using nearest-neighbor fill. Imputation leads to a large drop in all the metrics, indicating that our AFI module plays a critical role in helping the world model to extract meaningful dynamics from sparse and irregular EHR data. We then prove the importance of two phases training by comparing it with two variants: training only on phase 1 \textit{MDP1 Only} (no full imagined trajectory training) and training directly on phase 2 \textit{MDP2 Only}(without pretraining on real data). There is a clear improvement in all the metrics when moving from phase 1 to phase 2, simply training on partially imagined trajectories is not sufficient to fully explore the data. Directly training on full imagined data surprisingly shows good results, indicating that our world model is able to create meaningful trajectories that are beneficial for policy learning. 
To prove the validity of two-phase training, we also test on end-to-end (E2E) training, which shows higher WIS and remains low for the other two metrics.

(2) \textit{Variants of Imagination}: We experimented with two hybrid training variants: in phase 1, we used a traditional Actor Critic to train entirely on real data; in phase 2, we trained using a mix of real and imagined data. This setup is not as good as directly using a mixture of data in phase 1, this observation shows it necessary to have an early exposure to imagined data for policy training. We also tried a dynamic way of utilizing curriculum learning to gradually increase the percentage of imagined data, however, this seems overly unstable for training and yields a low WPDIS. 

(3)\textit{Choice of Imagined Length}: We evaluated the impact of varying the imagined to total steps ratio ($\tau=5, 10, 30, 50$) which corresponds to 10\%, 0.2\%, 0.6\%, and 0.99\% of the trajectory length. Best performance is observed with $\tau=10$, with decreasing performance with higher values. This suggests that a moderate number of imagined transitions is key to balancing realism and foresight, preventing drift from true dynamics.
\begin{figure}[h!]
    \centering
    \includegraphics[width=0.9\linewidth]{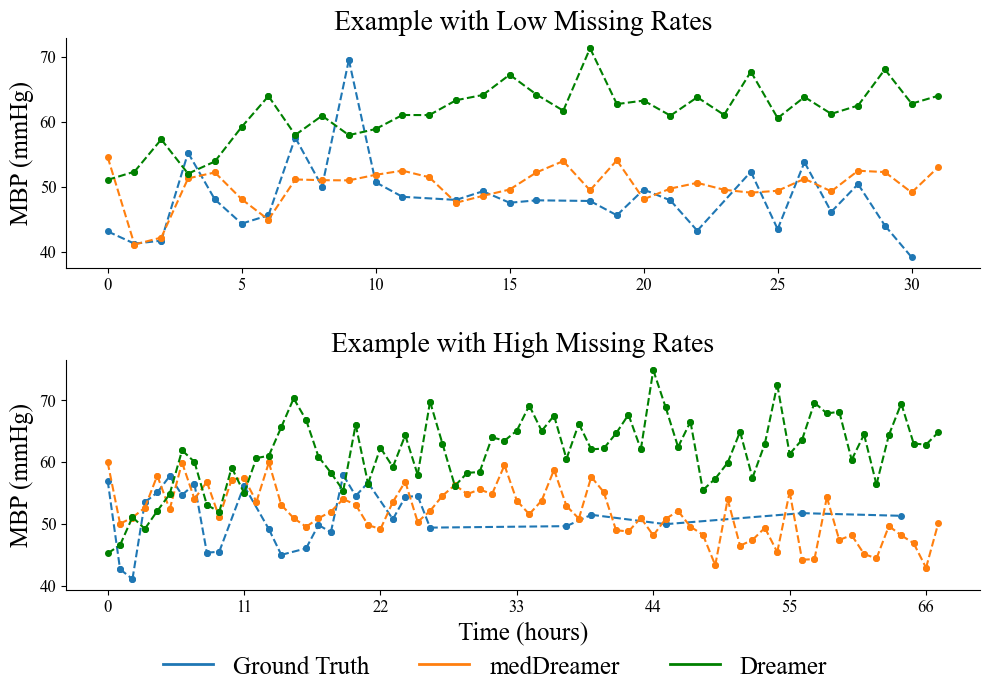} 
      \vspace{-2mm}
    \caption{Samples of mean blood pressure trajectories from original dataset, reconstruction of medDreamer and Dreamer with different missing rates.} 
    \label{fig:mbp} 
\end{figure}

\subsubsection{\textbf{RQ5: How reliable is the learned world model in capturing patient physiology and risk?}}
To evaluate the reliability of the learned world model, we assess its ability to reconstruct key physiological signals and reflect clinically meaningful patterns that relate to patient outcomes. Specifically, we report the model’s capacity to reproduce individual trajectories of an important vital sign mean blood pressure (MBP) under different missing rates. Figure \ref{fig:mbp} compares the original measurements to the values reconstructed by medDreamer and Dreamer. Our model's reconstructed trajectories demonstrate strong temporal alignment with the ground truth, while Dreamer tends to predict values around the global mean. Especially when the trajectory has many missing values, our model can still maintain similar trends. This highlights the model's ability to simulate patient states in an internally consistent and physiologically plausible manner, which is critical for reliable long-horizon planning. 
\begin{figure}[htbp]
  \centering
  \includegraphics[width=0.8\linewidth]{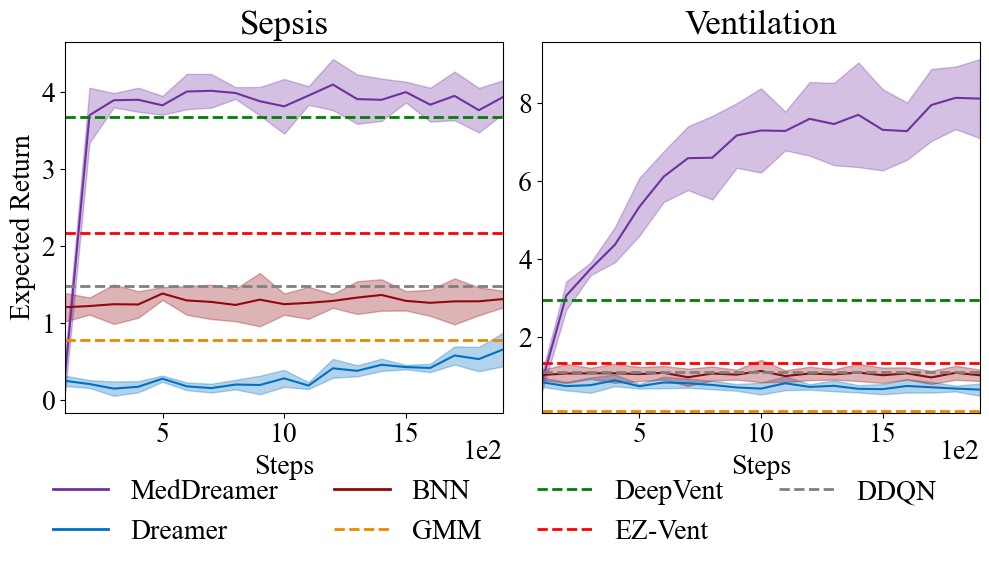} 
  \caption{Expected return over training steps.}
  \vspace{-2mm}
  \label{fig:ope}
\end{figure}
\subsubsection{\textbf{How does the policy's expected return evolve during Phases?}}
We plotted the expected return against training steps in Figure \ref{fig:ope}, it shows that when training progresses, the return increases steadily. medDreamer converges within 2000 epochs and consistently outperforms model-free and model-based baselines across both tasks, indicating better policy-alignment and higher returns. In contrast, other model-based baselines show limited improvement, and notably, Dreamer struggles to converge due to instability in joint end-to-end training.



\section{Conclusion}
We introduced medDreamer, a two-phase model-based RL framework for personalized treatment recommendation for irregular clinical data.  Extensive experiments on sepsis and ventilation tasks show that medDreamer effectively handles complex EHR data and considerably reduces estimated mortality, demonstrating its potential as a reliable clinical decision support system.
\bibliographystyle{plain}
\bibliography{references}
\newpage
\appendix
\section{Ethical Statement}
The collection of patient information and creation of the research resource in the MIMIC-IV database was reviewed by the Institutional Review Board at the Beth Israel Deaconess Medical Centre (number 2001-P-001699/14), which granted a waiver of informed consent and approved the data-sharing initiative. The eICU database has been approved by the Institutional Review Board of the Massachusetts Institute of Technology. After completing the National Institutes of Health’s online training course and the Protection of Human Research Participants Examination, we had the access to extract data from both the MIMIC-IV and the eICU databases. The study data are anonymous and deidentified. Our proposed method is designed solely to assist clinicians in their decision-making process by providing data-driven insights and recommendations. The final treatment decisions remain entirely under the control of the clinicians, ensuring that their expertise and judgment guide patient care. Importantly, all models were trained and tested using publicly available retrospective datasets, ensuring that no live patient data or clinical trials were involved in the development process. This approach aligns with established ethical standards, minimising potential risks and maintaining the highest level of transparency and safety.

\section{Features and Actions}
\label{sec:feature}
 Sepsis observations contain 40 physiological measurements, including demographics, vital signs, and laboratory test results, listed in Table \ref{table:mimic_features}. For the action space, two interventions are included: IV fluids and Vasopressors. We discretize the dosage of each treatment into four quantile bins along with a no-treatment option, resulting in five discrete levels per treatment. Consequently, the joint action space consists of \(5 \times 5 = 25\) possible treatment combinations.

\begin{table}[htbp]
\centering
\small
\caption{List of sepsis features.}
\begin{tabular}{p{2.5cm}p{5.5cm}}
\toprule
\textbf{Category} & \textbf{Feature Name} \\ \midrule
Demographics (5) &
  Age, Gender, Weight, Readmission, Elixhauser Score \\ \midrule
Vital signs (12) &
  Heart Rate, SBP, MBP, DBP, Respiratory rate, Temperature, SpO\textsubscript{2}, Shock Index, SOFA, GCS, FiO\textsubscript{2}, SIRS \\ \midrule
Lab values (22) &
  Lactate, PaO\textsubscript{2}, PaCO\textsubscript{2}, pH, Base excess, CO\textsubscript{2}, PaO\textsubscript{2}/FiO\textsubscript{2}, WBC, Platelet, BUN, Creatinine, PTT, PT, INR, AST, ALT, Total Bilirubin, Magnesium, Ionized Calcium, Calcium, Urine Output, Mechanical Ventilation \\ \midrule
Other (1) &
  Step ID \\ \bottomrule
\end{tabular}
\begin{flushleft}
Abbreviations - INR: International normalized ratio; PT: Prothrombin time; PTT: Partial thromboplastin time; SIRS: Systemic inflammatory response syndrome; WBC: White blood cell; GCS: Glasgow coma scale; SBP: Systolic blood pressure; MBP: Mean blood pressure; DBP: Diastolic blood pressure; SOFA: Sequential organ failure assessment; SpO\textsubscript{2}: Peripheral oxygen saturation; FiO\textsubscript{2}: Fraction of inspired oxygen; PaO\textsubscript{2}: Partial pressure of oxygen; PaCO\textsubscript{2}: Partial pressure of carbon dioxide; PaO\textsubscript{2}/FiO\textsubscript{2}: Arterial oxygen partial pressure to fractional inspired oxygen ratio; BUN: Blood urea nitrogen; AST: Aspartate aminotransferase; ALT: Alanine aminotransferase.
\end{flushleft}
\label{table:mimic_features}
\end{table}

For MV task, Table\ref{table:eicu_features} lists the 41 features selected for training. In our study, the action space is constructed as 18 possible discrete actions from combinations of low, medium, and high levels of the 3 ventilator settings: PEEP, FiO\textsubscript{2}, and ideal body weight–adjusted tidal volume (summarized in Table\ref{tab:mv_action}).
\begin{table}[h!]
\centering
\caption{Categorization of action levels for PEEP, FiO\textsubscript{2} and tidal volume.}
\begin{tabular}{ccc}
\toprule
Intervention & \textbf{Category} & \textbf{Threshold}\\
\midrule
\multirow{2}{*}{PEEP (cmH\textsubscript{2}O)} & Low    & $\leq 5$ \\
                                              & High   & $> 5$  \\
\midrule
\multirow{3}{*}{FiO\textsubscript{2} (\%)}    & Low    & $< 35$    \\
                                              & Medium & $35$–$50$   \\
                                              & High   & $\geq 50$  \\
\midrule
\multirow{3}{*}{Tidal Volume (ml/kg IBW)}     & Low    & $< 6.5$  \\
                                              & Medium & $6.5$–$8$  \\
                                              & High   & $\geq 8$  \\
\bottomrule
\label{tab:mv_action}
\end{tabular}
\end{table}

\begin{table}[htbp]
\centering
\small
\caption{List of MV features.}
\begin{tabular}{p{2.5cm}p{5.5cm}}
\hline
\textbf{Category} & \textbf{Feature Name} \\ \midrule
Demographics (4) &
  Age, Gender, Admission Weight, Elixhauser Score \\ \midrule
Vital signs (10) &
  Heart Rate, SBP, DBP, MBP, Respiratory Rate, Temperature, SpO\textsubscript{2}, SIRS, SOFA, GCS\\ \midrule
Lab values (27) &
   Richmond Agitation-Sedation Scale, Rate Std, Creatinine, Lactate, Bicarbonate, PaCO\textsubscript{2}, pH, Base Excess, Chloride, Potassium, Sodium, Glucose, Hemoglobin, Ionized Calcium, Calcium, PTT, PT, INR, Platelet, WBC, Albumin, Magnesium, Analgesic and Sedative Medication Administration, Neuromuscular Blocking Agent Administration, ETCO\textsubscript{2}, BUN, Urine Output\\
\bottomrule
\hline
\end{tabular}
\small
\label{table:eicu_features}
\end{table}


\section{Behavior Policy}
We created a behavior policy that imitates clinicians' actions using supervised learning. Specifically, we train and Long Short-Term Memory (LSTM) network with hidden size of 16 and hidden layer of 1 for both sepsis and MV tasks, where the learning rate is 0.0001 for sepsis task and 0.001 for MV. This behavior policy is utilized to calculate OPE metrics and serves as an approximate of clinician policy.

\section{Mortality vs Return}
\label{sec:mort_vs_return}
\begin{figure}[htbp]
  \centering
  \includegraphics[width=\linewidth]{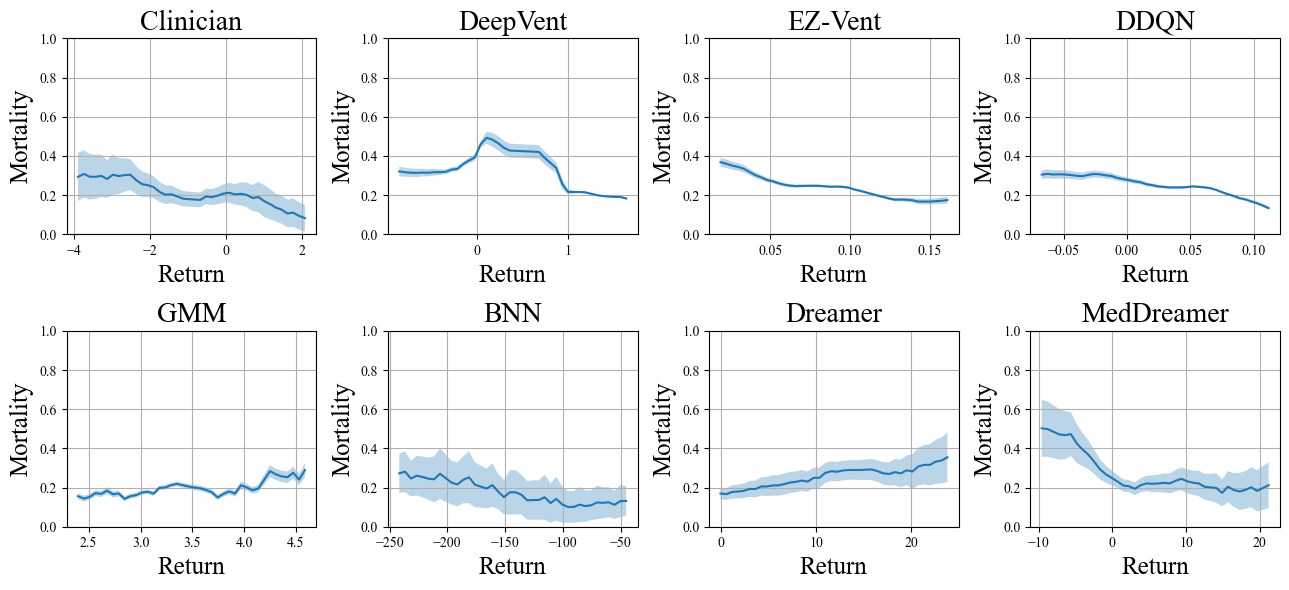} 
  \caption{Mortality vs return plot including baselines for sepsis task.}
  \label{fig:mort_vs_return_mimic}
\end{figure}

\begin{figure}[htbp]
  \centering
  \includegraphics[width=\linewidth]{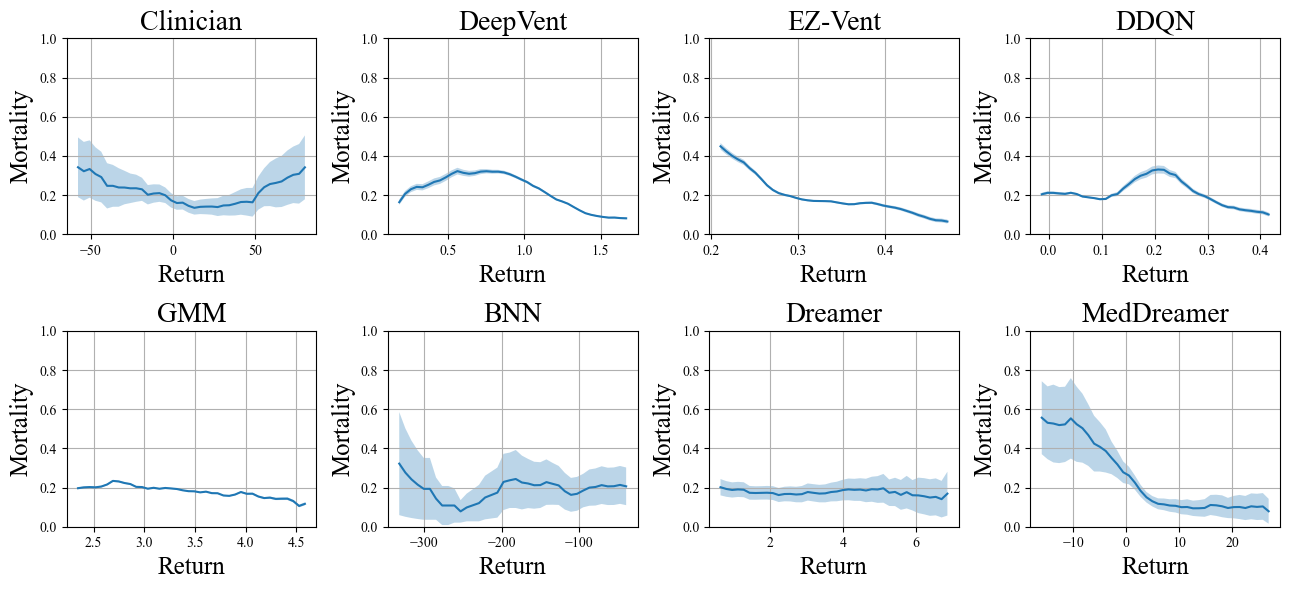} 
  \caption{Mortality vs return plot including baselines for MV task.}
  \label{fig:mort_vs_return_eicu}
\end{figure}

We analyze the relationship between expected return and predicted mortality across models for both the sepsis (Figure\ref{fig:mort_vs_return_mimic}) and MV (Figure\ref{fig:mort_vs_return_eicu}) tasks. Across both cohorts, medDreamer consistently demonstrates the expected strong inverse correlation between return and mortality—higher estimated returns correspond to lower mortality, suggesting that the learned reward signal effectively captures clinically meaningful outcomes. In contrast, Dreamer exhibits an unexpected upward trend in mortality with increasing returns in the sepsis task, indicating misalignment between its learned reward and actual patient outcomes. Model-free baselines such as DeepVent, EZ-Vent, and DDQN display generally flat or noisy trends, revealing limited correlation between return and mortality, and highlighting a lack of effective value shaping. Model-based EHR and BNN show moderate downward trends but with reduced magnitude and stability. Notably, clinician policies also show downward trends, validating the itermediate reward design. Overall, these findings reinforce medDreamer’s ability to learn a return function that more faithfully reflects outcome-relevant clinical improvements, leading to more interpretable and trustworthy decision-making.

\section{Additional results}
\subsection{Model Robustness Analysis}

The missingness observed in the dataset is intrinsic, largely representing informative missingness where the absence of a measurement conveys clinical significance. As the dataset lacks explicit labels to differentiate informative from random missingness, we evaluated model robustness by simulating increased data sparsity.We applied random masks to an additional 10\%, 20\%, and 50\% of the observed values. As demonstrated in Table \ref{tab:robustness}, the model retains strong performance on the Sepsis task with up to 20\% induced missingness.

\begin{table}[h]
    \centering
    \caption{Robustness experiment results on Sepsis task with varying mask percentages.}
    \label{tab:robustness}
    \begin{tabular}{lccc}
        \toprule
        \textbf{Mask\%} & \textbf{WIS} & \textbf{WPDIS} & \textbf{CWPDIS} \\
        \midrule
        0 & 9.24 & 0.27 & 0.19 \\
        10 & 9.23 & 0.25 & 0.19 \\
        20 & 9.17 & 0.25 & 0.19 \\
        50 & 8.44 & 0.22 & 0.15 \\
        \bottomrule
    \end{tabular}
\end{table}

\subsection{Evaluation on Renal Replacement Therapy}

To assess the generalizability of medDreamer across distinct patient populations and healthcare systems, we extended our evaluation to the AmsterdamUMCdb, a European ICU dataset. This analysis focused on the task of renal replacement therapy, where the agent is required to determine the necessity of dialysis and select the appropriate mode. It is important to note that the OPE metrics for this task are negative, reflecting a reward function design that incorporates a higher density of negative values compared to previous tasks. As presented in Table \ref{tab:amsterdam}, medDreamer consistently outperforms both the clinician and the baseline Dreamer model, validating the framework's adaptability to non-US datasets and distinct clinical decision-making tasks.

\begin{table}[h]
    \centering
    \caption{Performance comparison on AmsterdamUMCdb for the Renal Replacement Therapy task.}
    \label{tab:amsterdam}
    \begin{tabular}{lrrr}
        \toprule
        \textbf{Method} & \textbf{WIS} & \textbf{WPDIS} & \textbf{CWPDIS} \\
        \midrule
        Clinician & -0.67 & -0.08 & -0.03 \\
        dreamer & -0.82 & -0.07 & -0.03 \\
        medDreamer & -0.18 & -0.06 & -0.02 \\
        \bottomrule
    \end{tabular}
\end{table}

\subsection{Comparison with Additional Baselines}

We incorporated two additional baseline model results on the Sepsis dataset: DAG and MeDT. The comparative results are shown in Table \ref{tab:baselines}. Compared to DAG and MeDT, our method's designed AFI module endows our model with better performance on irregular temporal data. At the same time, our two-phase training strategy ensures safety under the clinical context. 

\begin{table}[h]
    \centering
    \caption{Comparison with additional baselines on the Sepsis dataset.}
    \label{tab:baselines}
    \begin{tabular}{lccc}
        \toprule
        \textbf{Method} & \textbf{WIS} & \textbf{WPDIS} & \textbf{CWPDIS} \\
        \midrule
        Random & 2.17 & -0.01 & -0.01 \\
        Clinician & 4.87 & 0.14 & 0.09 \\
        \midrule
        EZ-Vent\cite{liu2024reinforcement}(BCQ) & 10.13 & 0.02 & 0.07 \\
        DeepVent\cite{kondrup2023towards}(CQL) & \textbf{10.42} & 0.03 & 0.07 \\
        DDQN\cite{raghu2017deep} & 10.10 & 0.02 & 0.07 \\
        DAG\cite{yin2022deconfounding} & 6.12 & 0.03 & 0.01 \\
        MeDT\cite{rahman2024empowering} & 7.63 & 0.13 & 0.06 \\
        \midrule
        MBRL-BNN\cite{raghu2018model} & 4.90 & 0.00 & 0.01 \\
        MBRL-GMM\cite{li2022electronic} & -0.74 & -0.01 & -0.01 \\
        Dreamer\cite{hafner2024masteringdiversedomainsworld} & 6.85 & 0.20 & 0.13 \\
        \midrule
        medDreamer & 9.25 & \textbf{0.27} & \textbf{0.19} \\
        \bottomrule
    \end{tabular}
\end{table}

\section{Broader Impact}
\label{sec:broader_impact}

We believe that medDreamer is an important step toward applying reinforcement learning to personalized clinical decision-making, particularly for high-risk tasks such as sepsis treatment and mechanical ventilation management. By enabling policy learning in a structured latent space while relying on retrospective EHR data, our framework offers the potential to enhance treatment consistency, improve patient outcomes, and assist clinicians with reliable, data-driven support in an ethical and clinically grounded training approach.

The framework is designed to complement, not compete the clinicians. medDreamer's capability to effectively integrate both real and simulated patient trajectories offers an opportunity to uncover previously unexplored or underexplored, yet clinically driven scenarios, such as patterns that are difficult for human clinicians to detect, particularly in complex or rapidly evolving scenarios. As supported by the experimental results, we believe this could lead to improved patient outcomes, especially in resource-constrained environments or where standardization is lacking, such as in complex critical care. Additionally, these data-driven systems could serve as a foundation for broader research into decision support systems that combine clinical reliability with generalizability, offering guidance for future systems that adapt across diverse patient populations and care settings. The learned policies could even help explore previously unexplored treatment processes, helping advance clinical care research.

However, more comprehensive experimentation, scrutinized evaluations, and thoughtful deployment is needed. Similar with any AI-driven system, especial considerations are needed to ensure that the learned policies align with clinical standards, transparency of recommendations, and safeguards against biases in historical data. We believe medDreamer as a step towards safe, collaborative, and intelligent clinical support, which would enable more equitable and effective healthcare when responsibly deployed.

\section{Limitations}
\label{sec:limitations}
MedDreamr addresses key healthcare treatment recommendation challenges and presents several innovations to learn effective treatment policies as evident from the conducted evaluations. However, \textbf{similar to the clinical decision support systems in literature}, it operates under some assumptions and constrains as follows:

\begin{itemize}
    \item All models, including the baselines, are trained and evaluated on retrospective, observational data. Therefore the results rely on the fidelity of the learned reward function, and are limited by the absence of counterfactual ground truth.

    \item Performance of all the learned policies is assessed using estimated returns from a learned reward model. This provides an even ground for evaluation across all the tested models. Although we have followed the standard practice, this proxy measure may not perfectly capture true clinical utility.

    \item All experiments are conducted on MIMIC-IV and eICU datasets. medDreamer was able to generalise better across the tested tasks and showed superior generalizability among the models tested. However, the generalizability beyond the tested tasks, hospitals or patient populations needs to be established before real-world deployments.

    
    \item While medDreamer demonstrates improved estimated mortality, and close alignment to clinician policies in the past, additional assessments on clinician trustworthiness of the policy is needed with expert collaboration before real-world deployment.
\end{itemize}

Importantly, many of these limitations are not unique to medDreamer, rather showcase broader challenges in deploying reinforcement learning systems in healthcare. Despite these, medDreamer demonstrates strong performance, generalization, and clinical grounding. We see it as a principled and extensible foundation for future clinical decision support systems that prioritize reliability and real-world applicability.

\end{document}